\documentclass{article}

\usepackage{arxiv}

\usepackage[utf8]{inputenc} 
\usepackage[T1]{fontenc}    
\usepackage{hyperref}       
\usepackage{url}            
\usepackage{booktabs}       
\usepackage{amsfonts}       
\usepackage{nicefrac}       
\usepackage{microtype}      
\usepackage{lipsum}
\usepackage{graphicx}
\graphicspath{ {./images/} }

\newcommand{\ignore}[1]{}

\usepackage[linesnumbered,ruled]{algorithm2e}

\usepackage{soul, color,xcolor}
\usepackage{multicol}
\usepackage{amsmath}
\usepackage{booktabs}
 \usepackage{multirow}

\author{
Saba Sadeghi Ahouei\\
Optimisation and Logistics\\
School of Computer and Mathematical Sciences\\
The University of Adelaide\\
Adelaide, Australia
\And
Denis Antipov\\
LIP6\\
Sorbonne University\\
Paris, France
\And
Aneta Neumann\\
Optimisation and Logistics\\
School of Computer and Mathematical Sciences\\
The University of Adelaide\\
Adelaide, Australia
\And
Frank Neumann\\
Optimisation and Logistics\\
School of Computer and Mathematical Sciences\\
The University of Adelaide\\
Adelaide, Australia
}

\title{Feature-based Evolutionary Diversity Optimization of Discriminating Instances for Chance-constrained Optimization Problems}

\begin{document}

\maketitle

\begin{abstract}
Algorithm selection is crucial in the field of optimization, as no single algorithm performs perfectly across all types of optimization problems. Finding the best algorithm among a given set of algorithms for a given problem requires a detailed analysis of the problem's features. To do so, it is important to have a diverse set of benchmarking instances highlighting the difference in algorithms' performance. 
In this paper, we evolve diverse benchmarking instances for chance-constrained optimization problems that contain stochastic components characterized by their expected values and variances. These instances clearly differentiate the performance of two given algorithms, meaning they are easy to solve by one algorithm and hard to solve by the other.
We introduce a $(\mu+1)~EA$ for feature-based diversity optimization to evolve such differentiating instances. We study the chance-constrained maximum coverage problem with stochastic weights on the vertices as an example of chance-constrained optimization problems.
The experimental results demonstrate that our method successfully generates diverse instances based on different features while effectively distinguishing the performance between a pair of algorithms.

\end{abstract}

\section{Introduction}
Finding an algorithm that performs perfectly on all optimization problems is impossible. Hence, with the vast number of algorithms and heuristics developed over the years, it is crucial to determine suitable algorithms for solving a specific problem. To do so, it is effective to assess the behavior of different algorithms in solving the problem regarding its characteristics. A diverse set of discriminating instances that accentuate the difference in the performance of algorithms helps to study their behavior, strengths, and weaknesses in solving a particular type of problem~\cite{DBLP:conf/foga/BossekKN00T19}. An instance is called \emph{discriminating} for a pair of algorithms if it is easy to solve by one algorithm and hard to solve by the other. 

In this paper, we propose a feature-based diversity approach to evolve diverse sets of differentiating instances for chance-constrained optimization problems. Since most real-world problems involve uncertainty, investigating problems under uncertainty is one of the most critical aspects of the optimization field. 
Chance-constrained optimization is a method to tackle uncertainty, where constraints that involve stochastic components can be violated with a probability less than a small value of $\alpha$~\cite{charnes1959chance,DBLP:conf/gecco/XieHAN019}.

In our experiments, we consider the maximum coverage problem with a budget chance constraint as an example of chance-constrained problems, where the objective function is submodular. One of the characteristics of submodular functions that makes them relatable to many real-world applications, is diminishing returns. This means the benefit of adding elements to a solution set, does not increase with the growth of this solution set~\cite{DBLP:conf/ismp/Lovasz82,DBLP:journals/mor/NemhauserW78}.

\subsection{Related Work}

Submodular functions characterize subset selection problems where the marginal gain from adding an element to a solution decreases as more elements are included into this solution. 
Submodular functions have been widely studied in the literature~\cite{DBLP:journals/mor/NemhauserW78,DBLP:journals/mp/NemhauserWF78,vondrak1978submodularity,DBLP:books/cu/p/0001G14,DBLP:conf/icml/BianB0T17}. Recently, the optimization of monotone submodular functions under deterministic and chance-constrained knapsack constraints has been studied~\cite{DBLP:conf/gecco/NeumannB021,DBLP:conf/ppsn/NeumannXN22,DBLP:conf/gecco/PereraN24,DBLP:conf/gecco/Pathiranage0AN24a}. Chance-constrained optimization is a method to deal with uncertainty. Evolutionary algorithms have been shown to be effective in solving chance-constrained optimization problems~\cite{DBLP:conf/ppsn/NeumannN20,DBLP:conf/aaai/DoerrD0NS20,yan2024sliding,DBLP:journals/corr/abs-2303-01695}.

Benchmarking is an important aspect of optimization. It helps with the basic assessment of the performance of new algorithms, with assessing algorithms' behavior, and also with algorithm selection and configuration~\cite{DBLP:journals/corr/abs-2007-03488}. Analyzing the behavior of different algorithms on specific problems helps to identify the best algorithm to solve them. Diverse sets of differentiating instances are essential for evaluating algorithms' performance in solving a problem~\cite{DBLP:conf/foga/BossekKN00T19}. In~\cite{DBLP:journals/ec/GaoNN21} the authors proposed a new framework to evolve diverse travel salesperson problem (TSP) instances that are easy or hard to solve for a chosen heuristic. They studied the diversity of the features of these instances in multidimensional feature space. Moreover, this approach was adjusted to generate variations of a given image that are similar to it but differ in terms of selected image features~\cite{DBLP:conf/gecco/AlexanderKN17}. The experimental investigations in~\cite{DBLP:conf/gecco/NeumannG0019} demonstrate the performance of popular multi-objective indicators in evolving diverse sets of TSP instances and images according to various features. In~\cite{DBLP:conf/foga/BossekKN00T19} Bossek, et. al. introduced new creative mutation operators for evolving diverse sets of instances for TSP with multifaceted topologies. In~\cite{Saba:10.1145/3638529.3654181} the authors evolved reliable instances for chance-constrained submodular problems that are easy to solve for one algorithm and hard to solve for another by introducing a new method to calculate the performance ratio.

To the best of our knowledge, no study provides methods to generate diverse sets of discriminating instances for chance-constrained optimization problems in the literature. To address this gap, we introduce new feature-based evolutionary diversity algorithms to evolve such instances.

\subsection{Our Contribution}

We propose two feature-based evolutionary diversity optimization approaches to evolve diverse discriminating instances for chance-constrained optimization problems. We consider the average and standard deviation of both the expected values and the variances of stochastic elements in the chance constraint as the features (that is, four different features). In our experimental investigations, we use the chance-constrained maximum coverage problem with stochastic weights on the nodes. However, our proposed method can be used for any chance-constrained optimization problem where expected values and variances can quantify the stochastic components.
We first propose a new mutation operator designed to get diverse discriminating instances based on the average of expected values and variances as features. This method significantly improves the diversity of these features in the evolved instance sets compared to the initial population. 
Then,  we proposed another mutation operator to achieve greater diversity in the standard deviations of the expected values and variances. This operator focuses on increasing the diversity of these features within the set while maintaining the average unchanged in each instance. Our experimental investigations show that this operator has a great ability to improve the diversity of these features in the set. Our method can generate new benchmark sets which can be used as an excellent foundation for feature-based algorithm selection.

In Section~\ref{sec:section1}, we introduce the chance-constrained maximum coverage problem, the discounting performance ratio used to indicate how differentiating an instance is, and the diversity measure. In Section~\ref{sec:section2}, we introduce our feature-based evolutionary diversity optimization and features. 
In Section~\ref{sec:section3} and ~\ref{sec:section4}  we propose two novel mutation operators for independent and dependent features, respectively, each followed by experimental investigations on the chance-constrained maximum coverage problem. The results are then compared with those of the conventional method to evolve differentiating instances (with no diversity optimization).

\section{Preliminaries}\label{sec:section1}

In this paper, we evolve diverse sets of discriminating instances for chance-constrained optimization problems using feature-based diversity optimization.
We use the chance-constrained maximum coverage problem as an example of chance-constrained optimization problems for our experimental investigations. The objective of this problem is to cover as many vertices as possible from the graph, by picking a subset of the nodes, where each node covers itself and all adjacent nodes. In our setting a stochastic cost is associated with each node in the graph and the problem has a budget chance constraint. Formally, the problem is defined as follows. 
\begin{align}
        \text{Maximize} \qquad F(x)  =N(V'(x)), \\
        \text{Subject to} \qquad    Pr(C(x)>B)\leq\alpha,
        \label{eq:MC}    
\end{align}

\noindent where $V'(x)$ is the subset of nodes in a solution, $N(V'(x))$ is the number of the nodes that are covered in that solution, which are, the nodes in $V'(x)$ and all their neighbors, $C(x)$ is the cost function, and $B$ is the constraint budget. Here $\alpha$ is a small value that bounds the probability of going over the budget in the constraint.
In this problem, each node $i$, $1\leq i \leq n$ has a stochastic cost, characterized by expected value $\mu_i$ and variance $\sigma^2_i$. Since the costs of these nodes are independent, the expected cost and variance of a solution respectively are formulated as $\mu(x) = \sum_{i=1}^n \mu_i x_i$ and $\sigma^2(x) = \sum_{i=1}^n \sigma_i^2 x_i$. 

To address the chance constraint we use Chebyshev's inequality to estimate an upper bound for the probability of exceeding the constraint budget and reformulate constraint~\eqref{eq:MC} as
\begin{equation}\label{ch}
  \mu(x) + \sqrt{ \sigma^2(x)\cdot \frac{(1-\alpha)}{\alpha}} \leq  B.
\end{equation}

\noindent According to~\cite{DBLP:conf/gecco/XieHAN019}, solutions that satisfy the surrogate constraint~\ref{ch} also will satisfy the chance constraint $Pr(C(x)>B)\leq\alpha$.

We aim to evolve diverse instances that are easy to solve by one algorithm and hard to solve by another. To make sure the instances are differentiating, we need a measure that reflects the performance ratio of two algorithms for each instance.
For this purpose, we use discounting performance ratio (see Equation~\eqref{eq:PR}) proposed by~\cite{Saba:10.1145/3638529.3654181} which allows us to calculate the performance ratio while controlling the reliability of the instances in the chance-constrained setting. We are comparing the performance of heuristic algorithms, which are inherently stochastic. Since these algorithms produce different solutions with each run, to get more reliable instances we perform each pair of algorithms for $r$ independent runs on each instance and use $R^\prime(I)$ as the performance ratio.
\begin{equation}\label{eq:PR}
\begin{split}
        R(I) & =P(A_1^I)/P(A_2^I), \\ 
    R^\prime(I) & = Exp[R(I)] - K_\theta \cdot  std[R(I)],
    \end{split}
\end{equation}
where $P(A_1^I)$ and $P(A_2^I)$ are the best objective values that each algorithm $A_1$ and $A_2$ respectively can reach, and $R(I)$ determines the performance ratio of the pair for instance $I$. If any of the algorithms cannot find a feasible value before the stopping criteria is met, $P(A^I)$ would be set to a small value $\epsilon$.

$Exp[R(I)]$ and $ std[R(I)] $ are consequently the expected value and standard deviation of the performance ratios of instance $I$ in these $r$ runs, and $K_\theta$ is a constant value specified by the confidence level $\theta$. We use $R'(I)$ to measure the performance difference for our pair of algorithms to ensure all the offspring are differentiating. In this method by penalizing the expected value of the performance ratios by their standard deviation, we ensure the reliability of these chance-constrained instances.

To ensure the diversity of the features in a set, we use the feature-based diversity measure proposed by~\cite{DBLP:journals/ec/GaoNN21}. This measure indicates the contribution of each instance $I_i$ to the diversity of a population $P = \{I_1, I_2, ..., I_\mu\}$, which is related to the distance between the feature value of an individual and the feature values of the closest individuals in the population. Formally, if we enumerate the population in ascending order of the feature values $ft$, that is, for all $i \in [2..\mu - 1]$ we have $ft(I_1) \leq ft(I_i) \leq ft(I_{i+1}) \leq ft(I_\mu)$, then for these $i$ we define
\begin{equation}\label{eq:diversity}
     d(I_i, P )=(ft(I_i)-ft(I_{i-1})) \cdot (ft(I_{i+1})-ft(I_i)).
\end{equation}
For $ft(I_1)$ and $ft(I_\mu)$ we set the value of $d(I_i, P )$ to $+\infty$ so they are always kept in the population. If any individual has the same feature value as any other individual in the population we set their value of $d(I_i, X)$ to $0$. 

Furthermore, we calculate the diversity of a population by summing the diversity measures of all instances in the population, except for the instances with minimum and maximum feature values, since their diversity contribution is set to $+\infty$. Thus, the diversity of the population is formulated as follows
 \begin{equation}\label{set D}
     D_s(P) = \sum_{i = 2}^{\mu-1} d(I_i, P).
 \end{equation}

\section{Approach} \label{sec:section2}

\begin{algorithm}[t]
    Population $P \gets \mu$ individuals with performance ratio of at least $T$\;
    \While{stopping criteria is not met}
    {
        Calculate the feature value $ft$ for each individual\;
        Sort population $P = \{I_1, \dots, I_\mu\}$ in ascending order of $ft$, so that $ft(I_1) \leq ft(I_2) \leq  ... \leq ft(I_\mu)$\;
        Choose $I_j$ from $\{I_2, \dots, I_\mu\}$: with probability $\frac{1}{3}$ choose $I_1$, with probability $\frac{1}{3}$ chose $I_\mu$ and with probability $\frac{1}{3}$ choose any number from $\{I_2, \dots, I_{\mu-1}\}$ u.a.r.\;
        $I'_j \gets \textsc{mutate}(I_j)$\;
        \If{$R'(I_j) \ge T$}
        {
            Add $I$ to $P$, preserving the ascending order of $ft$\;
            Recalculate contributions of individuals $d(I_i, P)$\;
            Remove from $P$ an individual with the minimum $d(I_i, P)$, breaking ties u.a.r.\;
        }
    }
    \caption{The $(\mu+1)~EA_D$.}
    \label{alg:main}
\end{algorithm}

We develop a ($\mu + 1$)~evolutionary algorithm to generate diverse discriminating instances for the maximum coverage problem which we call the $(\mu + 1)$~EA$_D$ (see Algorithm~\ref{alg:main}). This algorithm starts with a population containing $\mu$ discriminating instances that meet a threshold $T$ of minimum performance ratio. As mentioned before, we are using the discounting performance measure to determine the difference in performance between a pair of algorithms for solving the problem at hand.
In each iteration, we choose one individual and alter the costs' expectation or variation of the nodes for that individual to get new offspring. By doing this, we are evolving instances with new constraints while preserving the graph's topology. If this offspring meets the minimum threshold of performance ratio we keep it in the population and remove an individual with the smallest contribution to the diversity. Otherwise, we continue to the next iteration.
The features we investigate in this study are the average and standard deviation of expected costs and variances in the graph. We aim to achieve population diversity by focusing on each feature separately. By optimizing diversity across these individual dimensions, we can better explore the solution space and the range of these features. These four features are formulated as follows

\begin{equation}\label{ft1}
        ft_1(I) =\mu^I_{ave}=\frac{1}{n} \cdot \sum_{i=1}^n \mu^I_i,
        \end{equation}

\begin{equation}\label{ft2}
        ft_2(I) =\sigma^I_{ave}=\frac{1}{n} \cdot \sum_{i=1}^n \sigma^I_i,
        \end{equation}

\begin{equation}\label{ft3}
        ft_3(I) =\sqrt{\frac{1}{n} \cdot \sum_{i=1}^n (\mu_i^I - \mu^I_{ave})^2} ,
\end{equation}

\begin{equation}\label{ft4}
        ft_4(I) =\sqrt{\frac{1}{n} \cdot \sum_{i=1}^n (\sigma_i^I - \sigma^I_{ave})^2} ,
\end{equation}
where $\mu^I_i$ and $\sigma^I_i$ are respectively the expected cost and variance of node $i$ in individual $I$.

 In Sections~\ref{sec:section3} and ~\ref{sec:section4} we introduce two novel mutation operators to improve the diversity of a set with respect to these features. These operators show a great ability to develop offspring with a significant contribution to the population's diversity while maintaining a minimum quality of performance difference.
 
  The idea behind the mutation operators is to reduce the minimum values and increase the maximum values of the features in the set of instances and try to make the values of $(ft(I_i)-ft(I_{i-1}))$ and $(ft(I_{i+1})-ft(I_i))$ equal in the other individuals. According to the diversity measure (see Equation~\ref{eq:diversity}), this maximizes the value of $(ft(I_i)-ft(I_{i-1})) \cdot (ft(I_{i+1})-ft(I_i))$  which is the value that shows the contribution of each solution to the set. This approach evolves differentiating instances with increasing contribution to the diversity, resulting in a final population with a diverse range of features.

 The fitness function of this algorithm is the contribution of each instance to the diversity of the population $d(I_i, P)$. Note that the fitness of individuals in the population depends on the whole population, and for the same individual, it might change from iteration to iteration.
 We use the elitist method for selection, which means that in every iteration the individual with the smallest contribution to the diversity is eliminated, with ties broken uniformly at random.
 The following sections provide a detailed explanation of these new mutation operators and experimental investigations.

\section{Evolving Discriminating Instances for Chance Constrained Problems with Diverse Features}
\label{sec:section3}
In the first mutation operator that we develop (see Algorithm~\ref{alg:mutation1}), we aim to expand the range of the average values for expected value and variances of costs, as the features (see Equations~\eqref{ft1} and~\eqref{ft2}). In the rest of this study, we refer to these features as feature~1 ($ft_1$) and feature~2 ($ft_1$).

\begin{algorithm}[t]
\For{each node $i \in \{1, \ldots, n\}$}{
    Choose $\delta_1^i$ according to normal distribution $N(0, \sigma_1)$\;
    $\mu_i \gets \min\{\mu_{max},~\max\{0,~\mu_i + Ind(j) \cdot |\delta_1^i|\}\}$\;
}
\caption{Mutation for an independent feature, demonstrated for expected values for instance $I_j$ (this works similarly for variances).} 
\label{alg:mutation1}
\end{algorithm}

In this algorithm, first, we sort the population by their feature value. Then we choose one individual as a parent. The parent is either the individual with minimum or maximum feature value or something in between, each with a probability $P_m=1/3$. All instances are defined by two vectors representing the expected values ($\mu_i$) and variances ($\sigma_i^2$) of their costs, where $i = \{1,\ldots,n\}$ and $n$ is the number of nodes in that instance's graph. There is a value associated with the selected parent, which indicates whether the feature of this instance should be increased or decreased to have the maximum contribution to the diversity of the set. To decide the value of this indicator we have
$Ind(j)=1$ if $(j=\mu \vee (j \not = 1 \wedge (ft(I_j)-(ft(I_{j-1})) \leq (ft(I_{j+1})-ft(I_j))$,
and $Ind(j) =-1$ otherwise.
Using this operator to mutate the individuals leads to an increase in the range of the feature within the population. For individuals $I_i$, $i \not \in \{1, \mu\}$ we change their weights or variances so that the difference in the feature value to $I_{i + 1}$ becomes closer to the difference in feature value to $I_{i - 1}$. This implies that feature values get balanced out among individuals that do not take on the maximum or minimum feature value.

For each node $i$ of the mutated individual $I_j$, we chose a random number $\delta_i \sim N(0, \sigma_1)$, where $\sigma_1$ is a parameter of the algorithm which can be used to tune the mutation strength, and then we add $(Ind(j)\cdot|\delta_{1}^i|)$ to the expected weight or variance of the node.
This novel mutation demonstrates a strong capability to evolve discriminating instances with diverse expected costs.

\begin{table}[t]

\small
\caption{Feature values of the diverse differentiating sets before (conventional) and after ($ (\mu+1)~EA_D$) using the feature-based diversity optimization method based on the independent features.}\label{table1}
\centering
\renewcommand{\arraystretch}{0.8}
\renewcommand{\tabcolsep}{4pt}
\resizebox{\textwidth}{!}{\begin{tabular}{@{}cccccccccccc@{}}
\toprule
\multirow{2}{*}{Graph}         & \multirow{2}{*}{\shortstack{ Graph \\Size}} & \multirow{2}{*}{instance set} & \multirow{2}{*}{feature} & \multicolumn{4}{c}{EA/FGA}         & \multicolumn{4}{c}{EA/GHC}          \\ \cmidrule(l){5-12} 
                               &                                     &                               &                          & Average & Std    & Min    & Max     & Average & Std    & Min    & Max    \\ \midrule
\multirow{8}{*}{lp-recipe}     & \multirow{8}{*}{205}                & \multirow{4}{*}{conventional} & $ft_1$                 & 497.40  & 19.77  & 455.55 & 530.09 & 503.43 & 23.40  & 467.80  & 563.94  \\
                               &                                     &                               & $ft_2$                 & 286.48  & 10.01  & 263.73 & 303.36 & 288.16 & 11.90  & 263.99  & 314.90  \\

                                &                                     &                               & $ft_3$  &82170.22	&5673.14	&69551.80	&92029.10	&83176.23	&6846.99	&69693.20	&99160.90\\
                                &                                     &                               & $ft_4$ &92502.18	&4414.19	&82550.00	&100310.00	&92920.87	&3455.68	&86684.80	&98544.90 \\
\cmidrule(l){3-12} 

                               &                                     & \multirow{4}{*}{$ (\mu+1)~EA_D$}        & $ft_1$                 & 590.63  & 109.38 & 438.70 & 851.40 & 499.29 & 37.29  & 439.01  & 563.94  \\
                               &                                     &                               & $ft_2$                 & 293.35  & 12.38  & 262.03 & 313.74 & 285.21 & 15.51  & 264.11  & 314.90  \\ 

                                &                                     &                               & $ft_3$  &81601.93	&6792.91	&68578.50	&92469.40	&73517.31	&21195.66	&5689.40	&99245.70\\
                                &                                     &                               & $ft_4$  &91725.23	&4862.35	&82557.00	&100315.00	&86562.02	&16676.04	&19999.10	&98544.60\\

\midrule
\multirow{8}{*}{ca-netscience} & \multirow{8}{*}{379}                & \multirow{2}{*}{conventional} & $ft_1$                 & 501.39  & 12.10  & 477.04 & 526.29 & 497.32 & 16.77  & 463.79  & 532.13  \\
                               &                                     &                               & $ft_2$                 & 288.66  & 8.25   & 276.34 & 303.10 & 285.42 & 9.23   & 270.86  & 302.42  \\ 
                               
                                &                                     &                               & $ft_3$ &83389.89	&4781.34	&76363.10	&91867.80	&81552.06	&5266.77	&73363.50	&91455.60 \\
                                &                                     &                               & $ft_4$  &93441.65	&3877.30	&88127.50	&100296.00	&92303.41	&3329.82	&86861.60	&97733.80  \\

                               \cmidrule(l){3-12} 
                               &                                     & \multirow{2}{*}{$ (\mu+1)~EA_D$}        & $ft_1$                 & 616.71  & 128.63 & 428.75 & 901.78 & 648.59 & 142.41 & 461.43  & 1000.00 \\
                                &                                     &                               & $ft_2$                 & 304.57  & 15.26  & 270.63 & 330.67 & 307.97 & 18.85  & 270.76  & 360.58  \\ 
                                &                                     &                               & $ft_3$ &83525.52	&7197.99	&70266.70	&97774.60	&87193.34	&11755.86	&73142.40	&117415.00 \\
                                &                                     &                               & $ft_4$  &94248.12	&4112.99	&88128.90	&100295.00	&93093.14	&3434.42	&87321.90	&98506.00 \\

                               \midrule
\multirow{8}{*}{impcol-d}      & \multirow{8}{*}{435}                & \multirow{4}{*}{conventional} & $ft_1$                 & 503.75  & 11.98  & 476.54 & 533.78 & 502.45 & 11.22  & 481.40  & 519.73  \\
                               &                                     &                               & $ft_2$                 & 289.10  & 6.73   & 275.74 & 305.03 & 288.21 & 7.62   & 273.52  & 300.09  \\ 
 &                                     &                               & $ft_3$ &83623.91	&3901.93	&76033.60	&93041.70	&83122.98	&4379.62	&74815.40	&90054.50\\
 &                                     &                               & $ft_4$ &93316.19	&2310.33	&88283.70	&98653.50	&92923.09	&3343.28	&87396.60	&99784.90\\

                               \cmidrule(l){3-12} 
                               &                                     & \multirow{4}{*}{$ (\mu+1)~EA_D$}        & $ft_1$                 & 636.11  & 138.97 & 436.56 & 951.66 & 703.55 & 156.71 & 469.30  & 1000.00 \\
                               &                                     &                               & $ft_2$                 & 306.77  & 17.75  & 270.50 & 339.62 & 320.29 & 51.85  & 272.58  & 537.77  \\ 
                                                               &                                     &                               & $ft_3$                 &82269.11 &7477.26 &66030.00	&97221.00	&86922.05	&12842.93	&67947.10	&117684.00\\
                              &                                     &                               &$ft_4$	&93306.45	&2625.77	&89068.30	&98670.40	&94427.28	&5461.48	&85711.20	&105159.00\\
                                                             
                               \midrule
\multirow{8}{*}{random graph}  & \multirow{8}{*}{500}                & \multirow{4}{*}{conventional} & $ft_1$                 & 500.06  & 13.59  & 481.50 & 526.34 & 502.09 & 15.22  & 477.09  & 534.52  \\
                               &                                     &                               & $ft_2$                 & 286.68  & 7.68   & 274.03 & 300.67 & 287.05 & 8.78   & 269.59  & 303.60  \\ 
                                &                                     &                               & $ft_3$ &82245.14	&4419.39	&75090.00	&90401.00	&82474.72	&5035.59	&72676.40	&92174.70 \\
                                &                                     &                               & $ft_4$  &92791.03	&2756.71	&87544.40	&97241.50	&91929.44	&3343.14	&84396.70	&96554.60  \\                             
                               
                               \cmidrule(l){3-12} 
                               &                                     & \multirow{4}{*}{$ (\mu+1)~EA_D$}        & $ft_1$                 & 500.06  & 13.59  & 481.50 & 526.34 & 542.59 & 44.97  & 470.11  & 625.91  \\
                               &                                     &                               & $ft_2$                 & 286.68  & 7.68   & 274.03 & 300.67 & 295.74 & 10.91  & 272.54  & 307.11  \\ 
                                &                                     &                               & $ft_3$ &82730.06	&4751.16	&75090.00	&90474.00	&75357.12	&21022.61	&763.78	&105918.00 \\
                                &                                     &                               & $ft_4$ &92848.15	&2808.63	&87544.40	&97241.50	&85536.89	&19544.34	&4383.95	&97952.90\\                                
                               
                               \midrule
\multirow{8}{*}{lp-agg}        & \multirow{8}{*}{615}                & \multirow{4}{*}{conventional} & $ft_1$                 & 495.50  & 9.25   & 471.88 & 510.73 & 498.33 & 10.87  & 475.78  & 514.13  \\
                               &                                     &                               & $ft_2$                 & 285.21  & 5.89   & 271.38 & 294.81 & 285.69 & 5.17   & 277.44  & 295.03  \\ 
                                 &                                     &                               & $ft_3$  &81379.80	&3337.23	&73649.40	&86913.40	&81646.60	&2952.12	&76975.50	&87040.50\\
                                &                                     &                               & $ft_4$  &92602.10	&2839.49	&85020.60	&98301.60	&91886.41	&2031.73	&88176.10	&95144.80\\                              
                               
                               \cmidrule(l){3-12} 
                               &                                     & \multirow{4}{*}{$ (\mu+1)~EA_D$}        & $ft_1$                 & 608.89  & 107.78 & 457.84 & 884.89 & 704.10 & 159.52 & 475.78  & 1000.00 \\
                               &                                     &                               & $ft_2$                 & 297.14  & 11.48  & 271.38 & 320.96 & 284.56 & 5.09   & 278.88  & 295.03  \\ 
                               
                                &                                     &                               & $ft_3$  &81466.54	&4164.77	&72998.30	&87869.50	&85433.61	&7245.04	&76756.10	&102854.00\\
                                &                                     &                               & $ft_4$ &92793.41	&3539.64	&84953.50	&98301.60	&93224.26	&2536.96	&88173.30	&96939.10    \\

                               \midrule
\multirow{8}{*}{can-715}       & \multirow{8}{*}{715}                & \multirow{4}{*}{conventional} & $ft_1$                 & 501.79  & 10.58  & 487.51 & 528.17 & 498.78 & 10.11  & 484.43  & 520.83  \\
                               &                                     &                               & $ft_2$                 & 288.17  & 5.45   & 279.27 & 300.66 & 286.10 & 6.17   & 277.77  & 300.41  \\ 
                                &                                     &                               & $ft_3$ &83069.60	&3148.01	&77990.60	&90398.50	&81893.84	&3559.84	&77158.70	&90244.00 \\
                                &                                     &                               & $ft_4$ &93329.39	&1562.15	&90158.20	&96657.70	&91737.04	&2663.46	&85955.40	&96008.10\\                                
                               
                               \cmidrule(l){3-12} 
                               &                                     & \multirow{4}{*}{$ (\mu+1)~EA_D$}        & $ft_1$                 & 610.07  & 92.73  & 478.04 & 795.74 & 632.91 & 127.79 & 467.77  & 923.97  \\
                               &                                     &                               & $ft_2$                 & 303.27  & 10.04  & 279.40 & 318.50 & 301.80 & 12.91  & 276.38  & 328.61  \\ 
                               
                                &                                     &                               & $ft_3$ &84137.51	&5801.29	&72738.00	&97244.00	&80107.78	&21471.62	&9216.42	&118738.00 \\
                                &                                     &                               & $ft_4$  &93312.95	&1729.50	&90158.80	&95554.90	&88606.75	&15626.40	&24665.10	&102811.00\\                               
                               \bottomrule
\end{tabular}}
\end{table}

\subsection{Experimental Setting}

In our experimental investigations, we generate diverse discriminating instances for the chance-constrained maximum coverage problem.
We start our $(\mu +1)$~EA$_D$ with a population of $20$ discriminating instances ($\mu = 20)$, with expected costs $\mu_i \in (0,\mu_{\max})$ and variances $\sigma_i^2 \in (0,\mu_i^2/3)$ in the chance constraint where $\mu_{\max} = 1000$. We calculate the cost budget as $B = n/30\cdot \mu_{\max}/2 $, where $n$ is the number of vertices of the graph, and we set the probability of exceeding the chance constraint budget $\alpha$, to $0.05$.
We use discriminating instances generated by the discounting performance measure with a $0.9$ confidence level ($\theta = 0.9$) from~\cite{Saba:10.1145/3638529.3654181}. We generate all $20$ instances independently from each other. For each graph, we use a percentage of $R^\prime(I)$ values for that instance as threshold $T$ for the performance ratio and make sure that all offspring added to the population throughout the optimization process meet this threshold. We calculate the threshold $T$ as $((0.80 \cdot (R^\prime(I) - 1)) + 1)$, which keeps all thresholds above $1$. By doing this, we keep the difference in performance between the two algorithms within a sufficient level while making the set more diverse in each iteration.

We are comparing the performance of the $(1+1)$ Evolutionary Algorithm (EA), with the Fast Genetic Algorithm (FGA), and the Greedy Hill Climber (GHC) in our experiments. Detailed explanations of these algorithms can be found in~\cite{DBLP:journals/asc/DoerrYHWSB20}. We evolve instances that are easy to solve by the $(1 + 1)$~EA and hard to solve by the FGA, and easy for the $(1 + 1)$~EA and hard for the GHC, which we denote respectively by EA/FGA and EA/GHC in our experimental results. We run each algorithm for 10000 iterations and perform 10 independent runs for each instance and use the expected value and standard deviation of performance ratios in these runs to calculate the discounting performance ratio. If any of the algorithms can not reach a feasible solution in any of the runs, we set the value of $P(A)$ for that run to a small value $\epsilon = 10^{-2}$.

As mentioned earlier, when mutating an individual, depending on the feature, we change the expected value or variances of the costs by adding or subtracting a small value from all of its values.
This value is chosen randomly from a normal distribution $N(0,3)$ for expected costs and $N(0,100)$ for costs' variances. We chose these parameters by trying different parameter values in experiments and checking the results to see which value works best for our setting. We run the $(\mu+1)~EA_D$ for 10000 iterations and perform $10$ runs per each feature and each combination of tested algorithms. This gives us $10$ sets of discriminating instances with diverse expected costs, each containing $20$ individuals for each graph.

\begin{table}[t]
\small
\caption{$D_S$ values (Equation~\eqref{set D}) for discriminating instance sets evolved by conventional and $ (\mu+1)~EA_D$ methods regarding average of weights ($ft_1$) and average of variances  ($ft_2$).}\label{table:diversity1}
\resizebox{\columnwidth}{!}{
\centering
\renewcommand{\arraystretch}{0.95}
\renewcommand{\tabcolsep}{7pt}
\begin{tabular}{@{}cccccc@{}}
\toprule
\multirow{2}{*}{Graph} &\multirow{2}{*}{feature}& \multicolumn{2}{c}{EA/FGA}                  & \multicolumn{2}{c}{EA/GHC}                  \\ \cmidrule(l){3-6} 
                  &     & conventional & $ (\mu+1)~EA_D$ &  conventional & $ (\mu+1)~EA_D$      \\ \midrule

lp-recipe    &\multirow{4}{*}{$ft_1$}                                                & 323.68                          & 10,009.30                         & 268.5                           & 767.32                          \\
ca-netscience    &                                            & 115.05                          & 13,778.22                         & 141.85                          & 20651.44                         \\
impcol-d          &                                           & 80.68                           & 11,710.95                         & 69.11                           & 14168.5                           \\
random graph    &                                             & 111.04                          & 111.04                            & 199.73                          & 1224.66                           \\
lp-agg            &                                           & 86.68                           & 10,915.55                         & 98.51                           & 14344.46                          \\
can-715          &                                            & 71.11                           & 5,722.73                          & 81.64                           & 10734.53                          \\  \midrule
lp-recipe       &\multirow{4}{*}{$ft_2$}                                            &29,778,773.08                   & 33,665,433.00                     & 30,856,754.00                   & 472,852,600.00                    \\
ca-netscience    &                                         & 12,397,990.24                   & 32,682,513.00                     & 15,100,157.88                   & 139,137,998.00                    \\
impcol-d       &                                           &12,368,648.03                   & 55,450,328.00                     & 13,859,104.77                   & 161,954,086.00                    \\
random graph   &                                       &  6,323,541.05                    & 7,301,225.92                      & 20,240,043.20                   & 673,454,960.00                    \\
lp-agg         &                                  & 7,516,936.70                    & 10,111,003.00                     & 4,727,498.00                    & 50,201,773.00                     \\
can-715      &                                & 6,774,002.20                    & 28,190,477.00                     & 4,526,037.80                    & 772,828,690.00                    \\ \bottomrule
\end{tabular}
}
\end{table}

\subsection{Experimental Results}
 In this section, we show the results of our experiments. They demonstrate the effectiveness of the proposed mutation operator in increasing the diversity of evolved instances. We compare a set of 20 instances, randomly selected from the 10 sets generated using our diverse optimization approach, with 20 instances evolved individually using the $(1+1)$~EA introduced by~\cite{Saba:10.1145/3638529.3654181} (which we refer to as the conventional method) to ensure a fair comparison. We tested our algorithm on six graphs of varying sizes, ranging from 200 to 715 nodes. As shown in Table~\ref{table1}, in the instances evolved by the feature-based diversity optimization, the range of both the average of expected costs (feature~1) and the average of costs' variances (feature~2), are significantly larger than instances generated by the conventional method. However, the range of standard deviation of the expected value and variances of the costs do not improve noticeably. This is natural as we are shifting all cost values in one direction in each iteration, and this does not significantly change the standard deviations of those values. 
 Additionally, as demonstrated in Table~\ref{table1}, for all features the increased standard deviation of the feature values indicates that the diversity of the first feature has been successfully improved after using our method. Figures~\ref{fig:1D,feature1} and~\ref{fig:1D,feature2} show the coverage of both instance sets evolved by the conventional and the $ (\mu+1)~EA_D$ method over the feature space for feature~1 and feature~2 respectively. The instances generated by the conventional method are very clustered in the feature space for both features, despite evolving separately with different random initial solutions. However, the instances evolved using the proposed feature-based diversity optimization provide significantly greater coverage of the feature space. As Figure~\ref{fig:1D,feature1} shows, evolved instances tend to have larger averages of expected costs and cover the feature space more extensively in that direction. This is because if the costs of the vertices in an instance are too small, then the problem gets too easy to solve by any algorithm. In some instances with very small expected costs, you can even include all of the graph nodes in a solution without exceeding the budget, leaving no opportunity to find discriminating instances.

\begin{algorithm}[t]
\caption{Mutation for a dependent feature demonstrated for expected values of instance $I_j$ (this works similarly for variances).}
\label{alg:mutation2}
$W_l = \{i \in V \mid \mu_i \leq \mu_{ave}\}$, 
$W_g = V \setminus W_l$\;

Choose $m \sim Pois(\lambda)$\;
$K = \min(m + 1,~\text{size}(W_l),~\text{size}(W_g))$\;

\For{$v \in \{0, \ldots, K\}$}{
    choose $\delta_2^v \sim N(0, \sigma_2)$\;
    choose nodes $s \in W_l$ and $t \in W_g$ randomly\;
    $\mu_s \gets \min\{\mu_{max},~\max\{0,~\mu_s + Ind(j) \cdot |\delta_2^v|\}\}$\;
    $\mu_t \gets \min\{\mu_{max},~\max\{0,~\mu_t - Ind(j) \cdot |\delta_2^v|\}\}$\;
}
\end{algorithm}

\begin{figure*}[t]
\centering
\caption{Distribution of average of weights ($ft_1$) for 20 instances. The left box plots show the features of the initial population and the right ones show them after using feature-based diversity optimization.}\label{fig:1D,feature1}
\includegraphics[width=\textwidth]{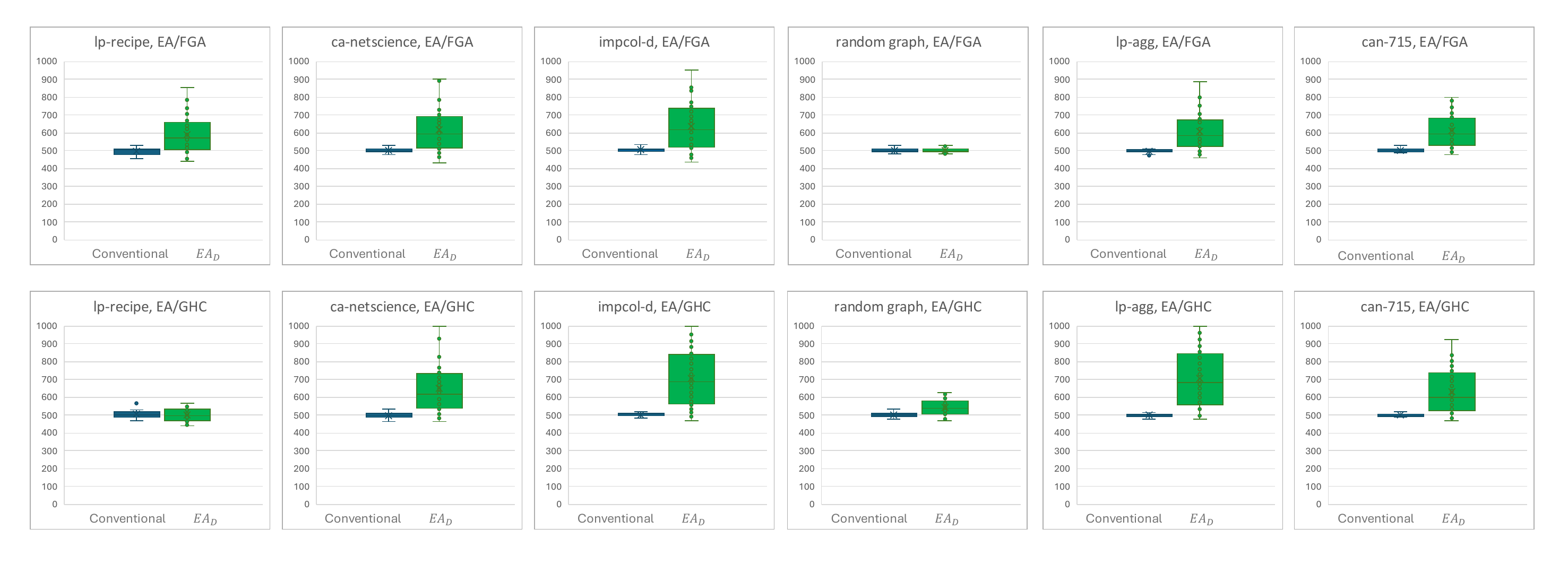}

\end{figure*}

\begin{figure*}[]
\vspace{-3mm}
\centering
\caption{Distribution of average of variances ($ft_2$) for 20 instances. The left box plots show the features of the initial population and the right ones show them after using feature-based diversity optimization.}\label{fig:1D,feature2}
\includegraphics[width=\textwidth]{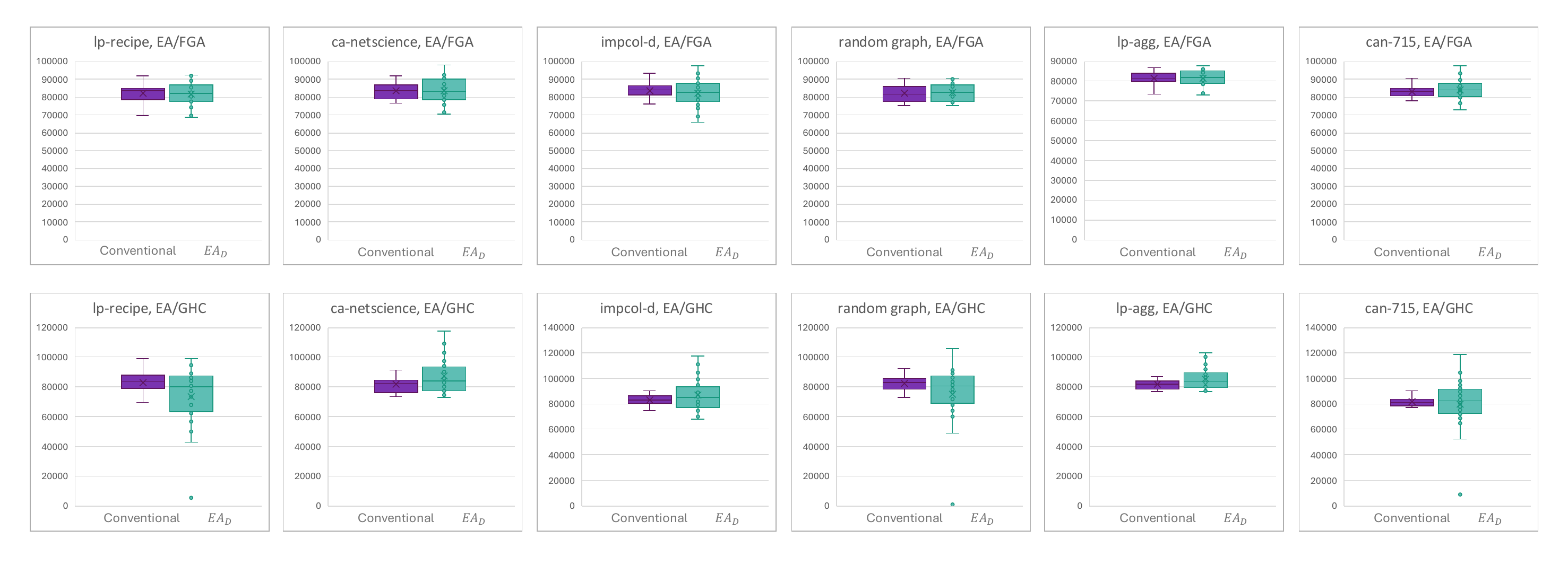}
\vspace{-5mm}

\end{figure*}

 The values of $D_s$ which represent a set's diversity, are shown in Table~\ref{table:diversity1}. As illustrated, these values are significantly larger for almost all graphs for the instance sets generated by $(\mu+1)~EA_D$, which is another indicator of high-quality instance sets in terms of diversity.
As stated earlier, the diversity in the standard deviation of expected values and variances of the costs (features~3 and~4), also do not provide sufficient coverage of the feature space (see Table~\ref{table1}). To address this, we propose a new mutation operator to diversify these features. We then proceed to demonstrate the strong performance of the proposed mutation operator through experimental investigations.

\section{Evolving Discriminating Chance Constraints with Diverse Dependent Features}
\label{sec:section4}

The mutation operator introduced in this section is tailored to evolve discriminating instances with diverse dependent features. We aim to increase the range of standard deviations of the expected values and variances of the costs as the features (respectively $ft_3$ and $ft_4$) while keeping the average of these values ($ft_1$, $ft_2$) unaffected. The core principle of this strategy is similar to the mutation proposed earlier. To indicate if we need to make these standard deviations smaller or larger, for the selected parent we use $Ind(j) \in \{-1, 1\}$. We decide on the value of $Ind(j)$ by considering the position of the instance compared to the other individuals in the population. If it has the maximum feature value we try to make the standard deviation larger by setting the value of $Ind(j)$ to $1$, if it has the minimum feature, we will do the opposite. For all other instances $I_i$ with $i \not \in \{1, \mu\}$, we determine the value of $Ind(j)$ in a way that brings the values of $(ft(I_i)-(ft(I_{i-1})) $ and $ (ft(I_{i+1})-ft(I_i))$ closer together.

 We alter the standard deviation of costs' expected values and variances without changing the averages. To do so we divide the nodes into two groups by looking into the respective averages for each feature. One group with values smaller, and the other with values larger than the average. By adjusting the expected costs or variances of these nodes to be closer to or further from the average by the same value, we can reduce or increase the standard deviation in that instance, without making any changes to the average. To modify the constraint in the mutation, we choose two nodes randomly, one from each category, and select a random normal value $\delta_2^i$, from $N(0, \sigma_2)$. We then add the value of $Ind(j)\cdot |\delta_2^i|$ to the expected cost of the individual chosen from the group with expected costs higher than average and discount the same value from the other one. By doing so, we are changing the standard deviation of the parent while keeping the average fixed. We do this $K$ times in the mutation operator for each iteration. To choose $K$, we first choose $m$ according to a Poisson distribution with parameter $\lambda$ and set $K = m+1$, to avoid iterations without mutating any nodes. The pseudocode of this mutation operator is shown in Algorithm~\ref{alg:mutation2}.

 This new mutation operator significantly enhances the diversity of standard deviations of the expected costs.

\begin{table}[t]
\small
\caption{$D_S$ values (Equation~\eqref{set D}) for discriminating instance sets evolved by conventional and $ (\mu+1)~EA_D$ methods regarding standard deviation of weights ($ft_3$) and  standard deviation of variances ($ft_4$).}\label{table:diversity2}
\resizebox{\columnwidth}{!}{
\renewcommand{\arraystretch}{0.95}
\renewcommand{\tabcolsep}{7pt}
\begin{tabular}{@{}cccccc@{}}
\toprule
\multirow{2}{*}{Graph} &\multirow{2}{*}{feature}& \multicolumn{2}{c}{EA/FGA}                  & \multicolumn{2}{c}{EA/GHC}                  \\ \cmidrule(l){3-6} 
                  &     & conventional & $ (\mu+1)~EA_D$ &  conventional & $ (\mu+1)~EA_D$      \\ \midrule
lp-recipe    & \multirow{4}{*}{$ft_3$}         & 79.55                & 1855.10              & 38.36                & 3752.07              \\ 
ca-netscience  &        & 32.78                & 495.67               & 15.66                & 4581.20              \\ 
impcol-d       &        & 17.06                & 2287.05              & 47.27                & 3898.64              \\ 
random graph  &         & 14.61                & 15.43                & 20.64                & 1474.18              \\ 
lp-agg        &         & 14.77                & 229.46               & 17.43                & 833.07               \\ 
can-715        &        & 9.69                 & 289.87               & 31.47                & 885.66  \\ \midrule

lp-recipe      & \multirow{4}{*}{$ft_4$}         & 14,000,313.97                 &   22,296,731.00                &  6,467,790.40                 &   22,605,636.00                \\ 
ca-netscience     &     & 8,872,710.77                &   14,888,695.00                &  5,515,740.80                &   57,315,042.00               \\ 
impcol-d      &         & 3,146,046.53                 &   20,687,811.00               &  9,499,351.87                 &  66,693,355.00                \\ 
random graph    &       &  5,174,800.50                 &   5,152,784.60                  &  4,436,745.78                 &   61,228,746.00                \\ 
lp-agg      &           &  8,444,281.95                 &   7,957,325.00                  &  3,069,856.78                 &    40,431,827.00               \\ 
can-715     &           & 1,348,041.36                 &  2,395,728.50                &  3,944,670.89                 &    61,146,158.00    \\ \bottomrule
\end{tabular}
}
\end{table}

\subsection{Experimental Setting}
This section illustrates the experimental investigations for evolving discriminating chance-constrained instances with diverse dependent features. Here features are the standard deviation of the expected values and variances of the costs in the constraint. We are performing feature-based diversity optimization based on these features while ensuring the average values of the expected values and variances of the costs do not change in the process. The problem setting remains the same as in the previous section. In each iteration, we change costs' expected values or variances (depending on the feature) of $2K$ items from the parent, where $K=m+1$, and $m$ is chosen randomly from the Poisson distribution $Pois(\lambda = 5)$. We are changing the expected costs of these selected nodes by random values which are chosen from the normal distribution $N(0,100)$ for expected costs and $N(0,3000)$ for variances. We run the algorithm for 10,000 function evaluations and perform 10 independent runs.

\begin{figure*}[t]
\centering
\caption{Distribution of standard deviation of weights ($ft_3$) for 20 instances. The left box plots show the features of the initial population and the right ones show them after using feature-based diversity optimization.}\label{fig:1D,feature3}

\includegraphics[width=\textwidth]{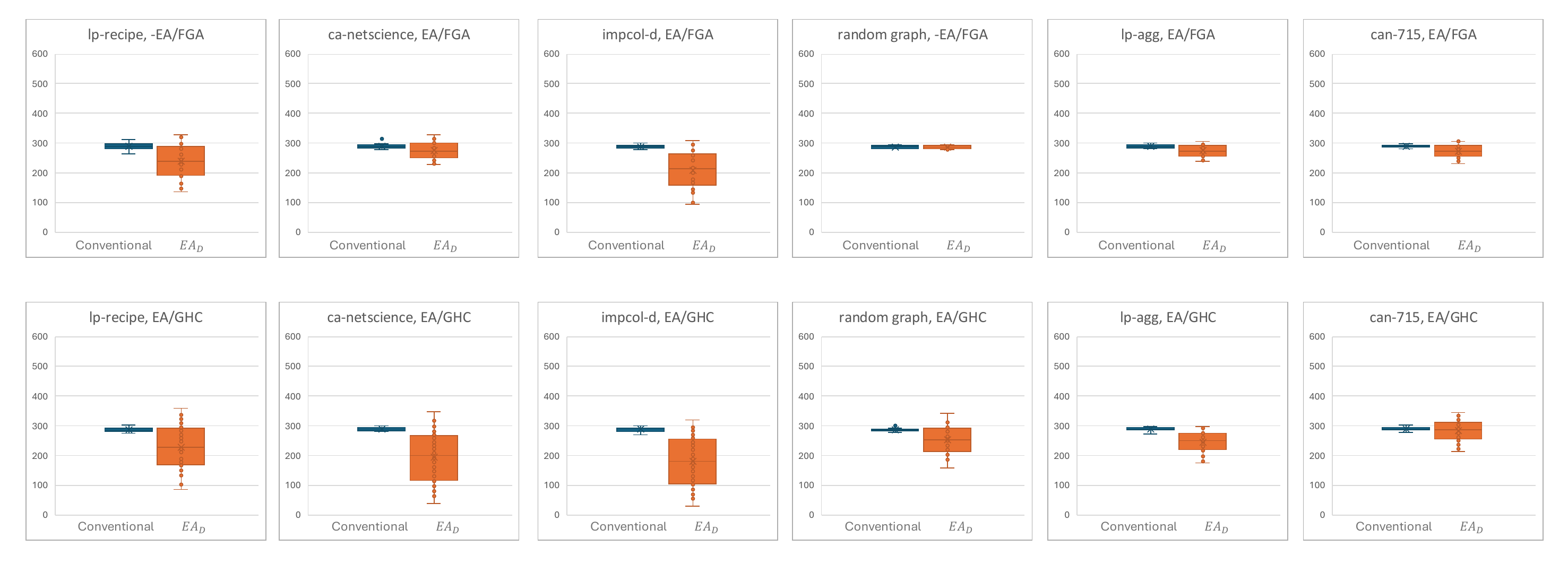}

\end{figure*}

\begin{figure*}[t]
\vspace{-2mm}
\centering
\caption{Distribution of standard deviation of variances ($ft_4$) for 20 instances. The left box plots show the features of the initial population and the right ones show them after using feature-based diversity optimization .}\label{fig:1D,feature4}
\includegraphics[width=\textwidth]{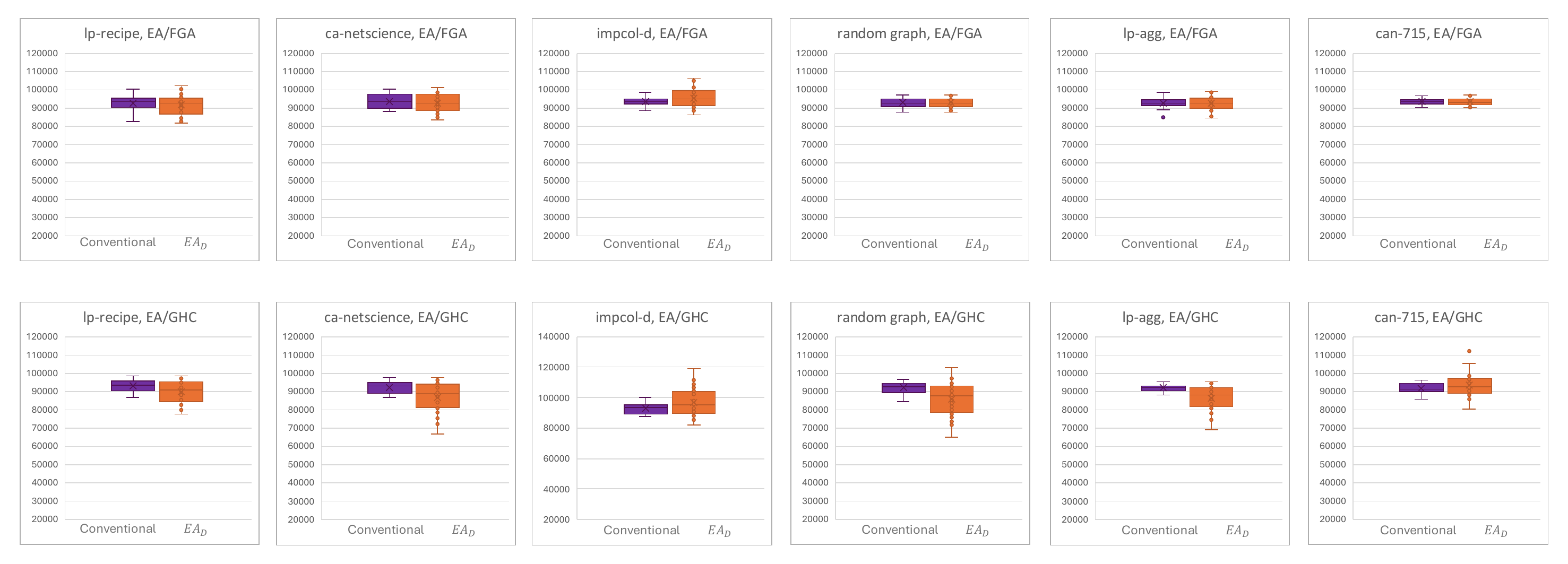}
\vspace{-8mm}
\end{figure*}

\subsection{Experimental Results}
This section presents the experimental results for instances generated by our feature-based diversity evolutionary algorithm regarding independent features. We start our algorithm with the same $20$ individuals as in the previous section. We run the algorithm 10 independent times and use one randomly to compare it to the initial set. The box plots showing the range of standard deviations of the expected values and variances of the costs before and after using our diversity optimization algorithm to evolve instances are represented in Figures~\ref{fig:1D,feature3} and~\ref{fig:1D,feature4} respectively. As the plot shows, the proposed $ (\mu+1)~EA_D$ is superior to the conventional method in terms of the diversity of the second feature. Obtaining diverse instances is more challenging for the last two problems with larger graph sizes. It is because we are doing the mutation with the same $\lambda$ value for all the problems despite the size of the problem. As a result, the number of elements altered during the mutation constitutes a smaller proportion of the total elements in the problem, leading to a smaller impact on the standard deviation. In all six graph problems, the range of $ft_3$ increases, mostly via decreasing the minimum values while the maximum values slightly improve. This is because raising the standard deviation of the expected costs too high creates an imbalance in the problem, with some nodes having very high costs and others very low. Consequently, the problem becomes too easy for both algorithms, as nodes with smaller expected costs are the obvious choices for covering the graph. This makes it more difficult to find discriminating instances with a performance ratio that meets the threshold, and consequently, it becomes harder to evolve diverse, discriminating instance sets. 

The values of $D_s$ for the instance sets generated by the conventional and feature-based diversity algorithms are presented in Table~\ref{table:diversity2}. These values, which reflect a set's diversity, are noticeably higher for all graphs after using our method except for "random graph" and "lp-agg" for $ft_4$. This can happen since $D_s$ is not the fitness function in our algorithm, thus we might be non-elitist in the optimization process regarding $D_s$.
One notable point is that achieving diversity in all of these features is challenging for the problem with the "random graph" when comparing the performance of $EA$ and $FGA$ (see Figures~\ref{fig:1D,feature1},~\ref{fig:1D,feature2},~\ref{fig:1D,feature3},~\ref{fig:1D,feature4}). This is because of the structure of this random graph which makes it an easier problem to solve and the high similarity of EA and FGA. In this random graph, the probability of an edge appearing is the same for all possible edges. This characteristic results in most nodes having very similar degrees, which makes the problem easier for many algorithms to solve.

\section{Conclusions}
In this paper, we proposed two feature-based diversity optimization approaches to increase the diversity of the average and standard deviation of expected values and variances of the costs for chance-constrained optimization problems. Focusing on single features in diversity optimization is important as it provides a clear perspective on the potential range of those features. Our methods demonstrate significant improvement in the range of these four features in the experimental investigation conducted on various graph instances. 
In future studies, it would be interesting to focus on increasing diversity by incorporating multiple features and expanding the range of features over multi-dimensional feature spaces.
\\
\\
\textbf{Acknowledgments:}
This work has been supported by the Australian Research Council (ARC) through grant FT200100536.

\bibliographystyle{unsrt}
\bibliography{main}
\end{document}